\documentclass[letterpaper]{article} % DO NOT CHANGE THIS
\usepackage{aaai25}  % DO NOT CHANGE THIS
\usepackage{amsmath}
\usepackage{amsfonts}
\usepackage{enumitem}
\usepackage{times}  % DO NOT CHANGE THIS
\usepackage{helvet}  % DO NOT CHANGE THIS
\usepackage{courier}  % DO NOT CHANGE THIS
\usepackage[hyphens]{url}  % DO NOT CHANGE THIS
\usepackage{graphicx} % DO NOT CHANGE THIS
\usepackage{natbib}  % DO NOT CHANGE THIS AND DO NOT ADD ANY OPTIONS TO IT
\usepackage{caption} % DO NOT CHANGE THIS AND DO NOT ADD ANY OPTIONS TO IT
\usepackage{algorithm}
\usepackage{algpseudocode}
\usepackage{booktabs}
\usepackage{array}
\usepackage{siunitx}
\usepackage{svg}
\usepackage{subcaption}

\usepackage{newfloat}
\usepackage{listings}
\usepackage{makecell}

\usepackage{xcolor}
\usepackage{tcolorbox}
\tcbuselibrary{skins}
\usepackage{hyperref}

\DeclareCaptionStyle{ruled}{labelfont=normalfont,labelsep=colon,strut=off} % DO NOT CHANGE THIS
\floatstyle{ruled}
\newfloat{listing}{tb}{lst}{}
\floatname{listing}{Listing}
\title{\textsc{Text2Zinc}: A Cross-Domain Dataset for Modeling Optimization and Satisfaction Problems in \textsc{MiniZinc}}
\author{
    Akash Singirikonda\textsuperscript{\rm 1},
    Serdar Kad{\i}o\u{g}lu\textsuperscript{\rm 1, 2} and Karthik Uppuluri\textsuperscript{\rm 2}
}
\affiliations{
    \textsuperscript{\rm 1}Department of Computer Science, Brown University, Providence, USA\\
        \textsuperscript{\rm 2}Artificial Intelligence Center of Excellence, Fidelity Investments, Boston, USA\\
        \{serdark@cs.brown.edu\}\\
}

\usepackage{bibentry}
\usepackage{color}

\makeatletter
\let\oldsubsubsection\subsubsection
\let\oldsection\section
\let\oldsubsection\subsection
\newenvironment{appendixformat}{%
   \onecolumn
   \appendix
   \setcounter{secnumdepth}{3}
   
   \setcounter{section}{0}
   \setcounter{subsection}{0}
   \setcounter{subsubsection}{0}

   \renewcommand{\section}{\@startsection{section}{1}{\z@}%
       {-2.0ex plus -0.5ex minus -.2ex}% space above
       {1.0ex plus .2ex}% space below
       {\Large\bf}}% format
   \renewcommand{\subsection}{\@startsection{subsection}{2}{\z@}%
       {-2.0ex plus -0.5ex minus -.2ex}% space above
       {1.0ex plus .2ex}% space below
       {\large\bf}}% format
   \renewcommand{\subsubsection}{\@startsection{subsubsection}{3}{\z@}%
       {-2.0ex plus -0.5ex minus -.2ex}% space above
       {1.0ex plus .2ex}% space below
       {\normalsize\bf}}% format
}{%
   \twocolumn
   \setcounter{secnumdepth}{0}
   \let\section\oldsection
   \let\subsection\oldsubsection
   \let\subsubsection\oldsubsubsection
}
\makeatother
\begin{document}
\nocopyright
\maketitle
\begin{abstract}
There is growing interest in utilizing large language models (LLMs) as co-pilots for combinatorial optimization and constraint programming tasks across various problems. This paper aims to advance this line of research by introducing \textsc{Text2Zinc}, a cross-domain dataset for capturing optimization and satisfaction problems specified in natural language text. Our work is distinguished from previous attempts by integrating \textit{both }satisfaction and optimization problems within a \textit{unified dataset} using a \textit{solver-agnostic} modeling language. To achieve this, we leverage  \textsc{MiniZinc}'s solver-and-paradigm-agnostic modeling capabilities to formulate these problems. Using the \textsc{Text2Zinc} dataset, we conduct comprehensive baseline experiments to compare execution and solution accuracy across several methods, including off-the-shelf prompting strategies, chain-of-thought reasoning, and a compositional approach.
Additionally, we explore the effectiveness of intermediary representations, specifically knowledge graphs. Our findings indicate that LLMs are not yet a push-button technology to model combinatorial problems from text. We hope that \textsc{Text2Zinc} serves as a valuable resource for researchers and practitioners to advance the field further.
\end{abstract}

\section{Introduction}
Generating constraint models from free-form natural language descriptions remains a significant challenge despite the groundbreaking advances in large language models (LLMs). Similarly, while constraint-solving techniques enjoy tremendous theoretical and practical advances, the cognitive barrier of translating problem descriptions into formal constraint models persists. This barrier is particularly acute, as domain experts who deeply understand their problem domain often lack the specialized knowledge required for formal modeling. The resulting dependency on modeling experts creates operational bottlenecks and can lead to misinterpretation of domain-specific requirements during the translation process.

High-level modeling languages such as \textsc{MiniZinc}~\cite{minizinc}, CPMpy~\cite{guns2019increasing}, and GAMS~\cite{Bussieck2004} have partially addressed these challenges by providing solver-agnostic approaches that are powerful and flexible. These modelling frameworks enable practitioners to focus on describing their problems without worrying about specific solution methods, making them especially useful for real-world applications, where requirements often change over time.

In parallel, LLMs still face significant challenges in handling the mathematical and logical reasoning needed for automated modeling. While language models are powerful at interfacing with natural text, they struggle with the consistency and precision required in declarative approaches, from basic type declarations to complex constraint relations. This gap between understanding textual descriptions and turning them into problem formulations indicates that more work is needed for automated modeling assistants (akin to code co-pilots).

\section{Our Contributions}
Our primary contribution is the introduction of \textsc{Text2Zinc}\footnote{\url{https://huggingface.co/datasets/skadio/text2zinc}}, a unified, cross-domain dataset that combines both optimization and satisfaction problems expressed in natural language. The dataset contains a diverse range of problem types across multiple domains, carefully curated from excellent resources including NL4OPT\footnote{\url{https://github.com/nl4opt/nl4opt-competition}}~\cite{nl4opt}, NLP4LP\footnote{\url{https://nlp4lp.vercel.app/}}~\cite{optimus}, ComplexOR\footnote{\url{https://github.com/xzymustbexzy/Chain-of-Experts}}~\cite{complexor}, CSPLib\footnote{\url{https://github.com/csplib/csplib}}~\cite{csplib}, and Hakank's collection\footnote{\url{http://www.hakank.org/ampl/}}~\cite{hakank}.
 We further enhance these problems through reformulation, metadata enrichment, and curation to ensure consistency and quality. This comprehensive collection provides a robust foundation for evaluating language-to-model translation.

The remainder of this paper is organized as follows. Section \hyperref[sec:background]{2} reviews fundamental concepts, including the distinction between optimization and satisfaction problems, an overview of \textsc{MiniZinc}, the role of knowledge graphs in combinatorial modelling, and relevant work utilizing LLMs for optimization. Section \hyperref[sec:dataset]{3} presents a detailed description of the \textsc{Text2Zinc} dataset. Section \hyperref[sec:methodology]{4} outlines our methodology, and Section \hyperref[sec:results]{5} discusses our results and analysis. Section \hyperref[sec:conclusion]{6} concludes with discussing our findings and future directions.

\section{Background}
\label{sec:background}

The following section provides background information about optimization and satisfaction problems, \textsc{MiniZinc} as a modeling language, and knowledge graphs as a representation structure.
\subsection{Optimization vs. Satisfaction Problems}

Optimization and satisfaction are two essential categories of decision-making problems commonly encountered in operations research, artificial intelligence, and mathematics. Although both require the satisfaction of constraints, their goals and approaches differ 

\paragraph{Optimization:} Optimization problems aim to find the \textbf{best solution} to a given problem by maximizing or minimizing an \textit{objective function} under a set of constraints. For instance, consider the following problem:

\textit{``An oil refinery manager has several million barrels of crude oil of different types allocated for production during the coming month. These resources can be used to make multiple different products. Each product has a price at which it sells. There are multiple production processes, each that uses some amount of each type of crude oil and produces some amount of each product. Each process has a cost per barrel of product produced. The crude oil has no separate cost as they have already been allocated. How many times should each process be executed to \textbf{maximize} the revenue for the next month?"}

In this example, the goal is to determine the optimal execution frequencies of the production processes to \textbf{maximize revenue}. Optimization problems have typically been solved by mathematical programming methods, which have roots in Linear Programming (LP). In this paradigm, it is common to utilize a reduction to a simpler inequality-constrained model, such as a linear program, which can be solved with highly developed methods that exploit its special structure~\cite{CPandOR}.

\paragraph{Satisfaction:} Satisfaction problems, in contrast, focus solely on finding a \textbf{feasible solution} that satisfies a set of constraints without optimizing a specific objective. A classic example of a satisfaction problem is the following:

\textit{
``Lecture timings need to be scheduled for courses across a limited number of periods. Each course requires a specific number of lectures and can only be assigned to certain periods due to availability constraints. Some courses have conflicts due to having common students and cannot be scheduled simultaneously. Additionally, there is a limited number of rooms that can be used and, thus, a maximum number of lectures that can occur simultaneously. How can we allocate lectures to periods while \textbf{ensuring all constraints are met?}"}

Here, the task is to satisfy the constraints dictated by the requirements to create a valid schedule. Satisfaction problems have typically been solved by methods in constraint programming, which has roots in logic programming. This method is commonly used for both verification and generation. This method often processes constraints independently, using domain reduction and propagation. \cite{CPandOR}.

\subsection{\textsc{MiniZinc}}
\textsc{MiniZinc}\footnote{\url{https://www.minizinc.org/}}~\cite{minizinc}  is a high-level constraint modeling language that supports both discrete and continuous optimization and satisfaction problems. Its solver-agnostic design allows communication with various solver backends. This enables leveraging different problem-solving paradigms, including Constraint Programming (CP), (Mixed) Integer Programming (MIP), and Boolean Satisfiability/Lazy Clause Generation (SAT). This flexibility is achieved through compilation to \textsc{FlatZinc}, an intermediate language that interfaces with different solvers, allowing the same \textsc{MiniZinc} model to be used across multiple backends without any code modifications. This enables end users to write the modeling code once and test different backends that best suit the task at hand. 

A key feature of \textsc{MiniZinc} is its use of global constraints, a concept originating in CP, which significantly simplifies the modeling process. For example, the all\_different($x_1, .. x_n$) constraint specifies that variables $x_1, x_2, ... x_n$ must take distinct values, replacing numerous pairwise inequality constraints that would be required in traditional MIP solvers. These global constraints can be specialized for particular solvers, often leading to improved performance~\cite{globalconstraints,minizinc}. Listing 1 shows the default expansion of the all\_different constraint from \textsc{MiniZinc} to \textsc{FlatZinc} as binary inequalities. If the solver is interfacing with Gecode, it can use Gecode's native all\_different constraint instead.

\begin{lstlisting}[numbers=none, captionpos=b, caption={All different constraint in MiniZinc.}]

% pairwise binary inequalities 
predicate all_different(array[int] of var float:x) =
forall(i,j in index_set(x) where i < j)(x[i] != x[j]); 

% built-in global constraint
predicate all_different(array[int] of var int:x) =
gecode_all_different(x); % native Gecode version for ints
\end{lstlisting}

Global constraints allow end-users to leverage intuitive higher-level declarative constraints rather than focusing on low-level decomposition.

\textsc{MiniZinc}'s practical utility for real-world applications is further enhanced by its clear separation of models (.mzn files) and instances (.dzn files). Models contain the problem structure, while instances provide specific input data, allowing a single model to be reused across multiple problem instances. The language structure is straightforward, consisting of four main components: decision variables, constraints, parameters, and an objective function for optimization problems or a satisfaction goal. This simple structure allows end users to create modular programs. Additionally, \textsc{MiniZinc} supports automated solution checking and model validation, which help evaluate model correctness during development.

Given its advanced features, we consider \textsc{MiniZinc} an excellent fit for bridging natural language and solver formulations. We hypothesize that higher-level language constructs, as offered by \textsc{MiniZinc}, serve as a good interface between LLMs and combinatorial formulations.

\subsection{Knowledge Graphs}

Knowledge Graphs (KGs) are structured representations of information where entities are represented as nodes and their relationships as edges\cite{knowledgegraph}. They have become fundamental tools in various domains, from powering Google's search engine to organizing scientific literature in academic databases(\cite{knowledgegraph}, \cite{dbpedia}, \cite{yago}, \cite{pubgraph}). In these graphs, real-world concepts and their relationships are captured in a format that both humans and machines can process effectively.
A knowledge graph is a particularly effective tool for representing information in a structured way, especially for combinatorial problems. 

In our context, parameters, variables, constraints, and objectives are depicted as nodes, interconnected through relationships that describe their interactions. For instance, an objective node is connected to all applicable constraints, highlighting the dependencies within the problem.
Generating code for less commonly known languages by large language models such as GPT-4 can be challenging when relying solely on an unstructured problem description and the input data. Instead, utilizing a structured knowledge graph as an intermediate representation can simplify this process. This graph models the problem at an atomic level, illustrating the intricate relationships between various components.
This structured representation, mainly when designed to imitate the logical structure of code, facilitates a smoother transition from unstructured problem description to executable code. It is crucial that this representation is ``code-inspired" to ensure it closely aligns with the final code structure, minimizing the gap between the model and this representation. We document both a knowledge graph generation prompt and an example knowledge graph in Appendix~\ref{appendix:knowledge_graph_generation}. 

\section{Related Work} % Related Work
A growing interest has been in leveraging large language models for optimization and constraint programming tasks. Early efforts like~\cite{nl4opt} focused on linear programming problems using entity recognition and logical forms, achieving promising results with ChatGPT (92.7\% accuracy on NL4OPT). Additionally, Holy Grail 2.0~\cite{tsouros2023holy} proposed a blueprint for building conversation modeling assistants. 
Significant improvements in execution accuracy of LLM generated \textsc{MiniZinc} code were achieved through using in-line annotation of entities in problem descriptions.~\cite{ner4opt2023, ner4opt2024}
 Sophisticated approaches emerged with a multi-agent Chain-of-Experts framework~\cite{complexor}, and LLM-agent, Optimus~\cite{optimus}, which developed a modular system for handling complex problem descriptions. However, these systems remain tied to specific solvers such as Gurobi and Cvxpy. The challenge of data scarcity was addressed by data augmentation, leveraging CodeT5 to achieve higher accuracy than zero-shot LLM approaches, although still limited to linear programming problems using PuLP~\cite{synthesis}.
Significant strides were made in training custom LLMs for optimization modeling through their OR-Instruct framework but remained constrained by single-solver dependency~\cite{orlm}. A notable contribution to evaluation methodologies came from the MAMO benchmark, focusing on LLMs' mathematical modeling processes rather than solution correctness. In CP, LLMs' potential in search space optimization has been demonstrated through generating streamliners using \textsc{MiniZinc}~\cite{streamllm}. Natural language to constraint model translation has also been explored through a simple decomposition-based prompting approach with GPT models~\cite{tsouros2023holy}. Building on this, in-context learning strategies such as Retrieval Augmented Generation (RAG) have been explored to build constraint models in CPMPY~\cite{michailidis_et_al:LIPIcs.CP.2024.20}. Domain-specific applications have also emerged, focusing on supply chain optimization while preserving data privacy~\cite{optiguide} and diagnosing infeasible optimization problems through interactive conversations~\cite{optichat}.

Our work, \textsc{Text2Zinc}, addresses several key limitations in existing research. First, unlike previous datasets focusing on single problem types, we uniquely integrate both satisfaction and optimization problems spanning continuous and discrete domains. Second, we leverage \textsc{MiniZinc}'s solver-agnostic modeling capabilities, moving beyond the solver-specific approaches of previous attempts. Finally, we explore the effectiveness of knowledge graphs as novel, structured, intermediate representations, an aspect unexplored in the existing literature.

\section{\textsc{Text2Zinc}}
\label{sec:dataset}
%\todo{look for serdar's problem in csplib}
The \textsc{Text2Zinc} dataset\footnote{\url{https://huggingface.co/datasets/skadio/text2zinc}} comprises of 110 carefully selected and augmented problem instances from five primary sources that blends optimization and satisfaction problems. 

\begin{itemize}
    \item NLP4LP~\cite{optimus}
    \item ComplexOR~\cite{complexor}
    \item LPWP~\cite{nl4opt} 
    \item CSPLib~\cite{csplib} 
    \item Hakank's collection~\cite{hakank}
\end{itemize}

\begin{figure*}[t]
    \centering
    \includegraphics[width=0.8\textwidth]{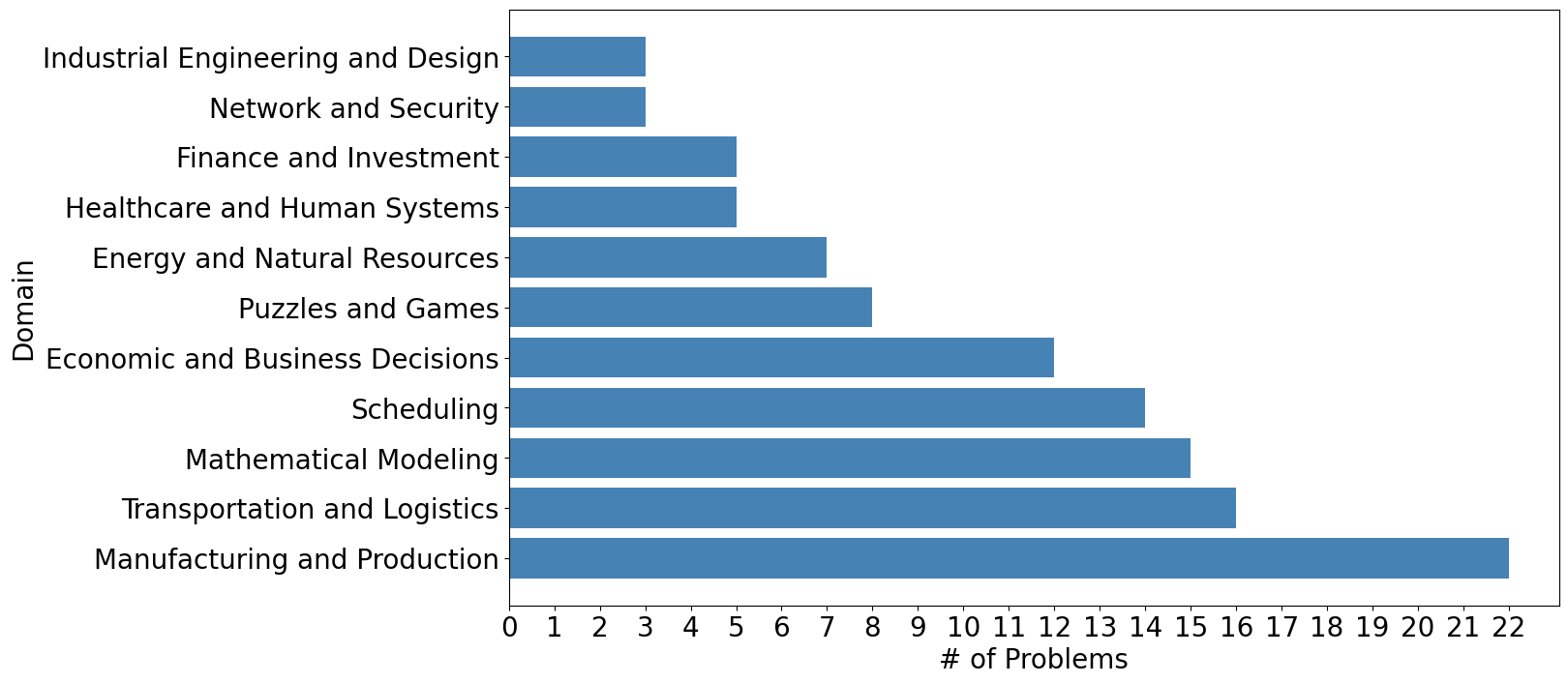}
    \caption{Distribution of problems across domains.}
    \label{fig:domains}
\end{figure*}

From this larger collection, we carefully curated problems that clearly distinguish between data and parameter components. We standardize the diverse input formats (json, dzn, mzn, html, txt) to our unification format. To serve as supervised labels, we provide ground truth outputs for most problems to enable validation. Each problem instance is enriched with extensive metadata (e.g., domain) to support diverse modeling approaches and future research needs. The dataset spans 11 application domains, as shown in Figure~\ref{fig:domains}. 

In the following, we detail the characteristics of each source and our specific contributions to their respective problems.

\begin{itemize}
    \item \textbf{NLP4LP: 65 Problems}~\cite{optimus} A collection of 67 LP, MILP, and MIP problems, originally represented in structured natural language. We converted the data from JSON to DZN format via a Python script and refined problem statements for clarity. Where included, we verified our generated code against the objective values provided in the dataset. We include 65 problems from this dataset.
    % This is a 67 problem dataset consisting of LP, MILP, and MIP problems. These problems were originally present in a structured natural language representation. Data files were converted from .json to .dzn files via python script. Generated code was verified against the objective values from the dataset. Problem statements were rephrased.
    %\item NL4OPT/LPWP: We randomly sampled 30 problems from the 1101 elementary-level linear programming (LP) problems. For all these problems, data files are not present. We use the exact text description on these problems. These problems do not have data files and all the information about the problems themselves is included in the descriptions. All problems are LP or MIP problems. The LPWP dataset (Ramamonjison et al., 2022b) is collected from the NL4Opt competition.
    \item \textbf{ComplexOR: 7 Problems}~\cite{complexor} A diverse collection of 37 operations research problems sourced from academic papers, textbooks, and industry applications, spanning domains such as supply chain optimization, scheduling, and warehouse logistics. We sampled seven sampled representative problems from this dataset.
    % These 37 problems cover a variety of operations research problems from diversified sources such as academic papers, textbooks, and real world industry scenarios, from domains such as supply chain optimization, scheduling, warehouse logistics, etc. For our dataset, we randomly sampled 7 problems from this dataset.
    \item \textbf{LPWP: 5 Problems}~\cite{nl4opt} This is a collection 1101 elementary-level linear programming (LP) problems collected from the Nl4Opt Competition~\cite{pmlr-v220-ramamonjison23a}. For these problems, we extracted parameters from problem descriptions into data files and modified problem descriptions accordingly.
    \item \textbf{CSPLib: 11 Problems}~\cite{csplib} A comprehensive library of 95 test problems for constraint solvers, featuring problems across various domains, including bin-packing, combinatorial mathematics, and scheduling. We included 11 problems from this collection and enhanced their existing \textsc{MiniZinc} solutions with detailed comments. 
    % CSPLib is a library of test problems for constraint solvers. The library consists of problems in a variety of areas, including bin-packing, combinatorial problems, and scheduling problems. Many problems have existing solutions in MiniZinc which we commented manually and via GPT4. 
    \item \textbf{Hakank's Collection: 22 Problems}~\cite{hakank} An extensive collection of over 1000 \textsc{MiniZinc} models spanning combinatorial problems, puzzles, operations research, and global constraints. We extracted problem features and descriptions from the extensively documented model files to standardize 22 problems.

\end{itemize}
\subsection{Dataset Format}
In our unified \textsc{Text2Zinc} format, each problem instance is divided into four components: \emph{Input}, \emph{Data}, \emph{Model}, and \emph{Output}. Detailed explanations of each component is as follows:

\medskip
\noindent \textbf{Input:} A JSON file that encapsulates a comprehensive description of a problem instance. This file is organized into several key sections:

\begin{itemize}[label=$\circ$]
    \item \textbf{Problem Description:} A standardized natural language problem description of the problem. To maintain abstraction, we avoid including specific parameter names and values, instead incorporating them in the data file. This retains a neutral language when describing the problem. 
    \item \textbf{Parameters:} Additional information about the parameters that are included in the \texttt{data.dzn} file. Each parameter is provided with a natural language explanation of its meaning and format, an associated symbol corresponding to its name in the \texttt{data.dzn} file, and a specification of its shape, which can be either an n-dimensional list or a scalar, represented by an empty list.
    \item \textbf{Output Specification:} A detailed explanation of the expected output format, including the name and shape of any decision variables to be output. This is required to evaluate satisfaction problems.
    \item \textbf{Metadata:} Additional contextual information, such as a descriptive problem title, a domain, and subdomain that enrich the problem, an objective (one of \texttt{satisfy}, \texttt{maximize}, or \texttt{minimize}), and a list of automatically generated keywords that reflect the constraints used in the problem, such as all\_different and $\leq$.
    
    \item \textbf{Unique Identifier:} A key linking the problem instance to its source. 
\end{itemize}

Figure~\ref{fig:input} presents an example input file.

\medskip
\noindent \textbf{Model:} A MZN file containing the \textsc{MiniZinc} model file that formulates the problem and serves as a verifier for satisfaction problems. The model can be used as a verifier by treating the output from the LLM-generated code in DZN format as input to the model. The constraints are decomposed where possible for better clarity. 

\medskip
Figure~\ref{fig:minizinc} shows an example model file.

\onecolumn
\begin{figure}[t]
    \centering
    \scalebox{0.90}
    {% Define colors
\definecolor{promptbg}{RGB}{220,240,255}
\definecolor{outputbg}{RGB}{255,250,220}
\definecolor{bordercolor}{RGB}{100,100,100}
\definecolor{headerbg}{RGB}{150,180,220}
\definecolor{headerfg}{RGB}{255,255,255}

\lstset{
    basicstyle=\ttfamily\small,
    columns=fullflexible,
    breaklines=true
}

% Custom Box Style
\tcbset{
    mybox2/.style={
        colframe=bordercolor,
        colback=promptbg,
        coltitle=headerfg,
        fonttitle=\bfseries,
        boxrule=0.5mm,
        width=17cm,
        rounded corners,
        enhanced,
        colbacktitle=headerbg,
        title style={left color=headerbg, right color=headerbg, rounded corners},
    },
    outputbox/.style={
        colframe=bordercolor,
        colback=outputbg,
        coltitle=headerfg,
        fonttitle=\bfseries,
        boxrule=0.5mm,
        width=16cm,
        rounded corners,
        enhanced,
        colbacktitle=headerbg,
        title style={left color=headerbg, right color=headerbg, rounded corners},
    }
}

\lstset{
    basicstyle=\ttfamily\small,
    columns=fullflexible,
    breaklines=true,
    tabsize=2, % Sets tab width to 2 spaces
}

\begin{tcolorbox}[mybox2, title=input.json]

    \begin{lstlisting}
{
  "parameters": [
    {
      "definition": "Number of courses",
      "symbol": "courses",
      "shape": []
    },
    {
      "definition": "Number of periods",
      "symbol": "periods",
      "shape": []
    },
    {
      "definition": "Number of rooms available",
      "symbol": "rooms",
      "shape": []
    },
    {
      "definition": "Binary matrix where A[i,j]=1 indicates lectures of course i can be scheduled at period j",
      "symbol": "available",
      "shape": ["courses", "periods"]
    },
    {
      "definition": "Conflict matrix where M[i,j]=1 if courses i and j have common students",
      "symbol": "conflict",
      "shape": ["courses", "courses"]
    },
    {
      "definition": "Array containing the number of lectures required per course",
      "symbol": "requirement",
      "shape": ["courses"]
    }
  ],
  "output": [
    {
      "definition": "Timetable grid where 1 represents a scheduled lecture and 0 represents an unscheduled lecture",
      "symbol": "timetable",
      "shape": ["courses", "periods"]
    }
  ],
  "description": "Lecture timings need to be scheduled for courses across a limited number of periods. Each course requires a specific number of lectures and can only be assigned to certain periods due to availability constraints. Some courses have conflicts due to having common students and cannot be scheduled at the same time. Additionally, there is a limited number of rooms that can be used and thus a maximum number of lectures that can occur simultaneously. How can we allocate lectures to periods while ensuring all constraints are met?",
  "identifier": "or_lp_ip_scheduling_problem_2",
  "metadata": {
    "name": "Timetable Problem", "domain": "Scheduling", "objective": "satisfy", "source": "hakank", "constraints": [
      "forall", "<=", "+", "=", "sum"]
  }
}


\end{lstlisting}

\end{tcolorbox}}
    \caption{An example input with description, parameters, metadata, and output fields.}
    \label{fig:input}
\end{figure}

\begin{figure}[t]
    \centering
    {% Define colors
\definecolor{promptbg}{RGB}{220,240,255}
\definecolor{outputbg}{RGB}{255,250,220}
\definecolor{bordercolor}{RGB}{100,100,100}
\definecolor{headerbg}{RGB}{150,180,220}
\definecolor{headerfg}{RGB}{255,255,255}

\lstset{
    basicstyle=\ttfamily\small,
    columns=fullflexible,
    breaklines=true
}

% Custom Box Style
\tcbset{
    mybox2/.style={
        colframe=bordercolor,
        colback=promptbg,
        coltitle=headerfg,
        fonttitle=\bfseries,
        boxrule=0.5mm,
        width=17cm,
        rounded corners,
        enhanced,
        colbacktitle=headerbg,
        title style={left color=headerbg, right color=headerbg, rounded corners},
    }
}

\lstset{
    basicstyle=\ttfamily\small,
    columns=fullflexible,
    breaklines=true,
    tabsize=2, % Sets tab width to 2 spaces
}

\begin{tcolorbox}[mybox2, title=model.mzn]

    \begin{lstlisting}
include "globals.mzn";

% Input parameters
int: courses;
int: periods;
int: rooms;

array[1..courses, 1..periods] of int: available;
array[1..courses, 1..courses] of int: conflict;
array[1..courses] of int: requirement;

% Decision variables
array[1..courses, 1..periods] of var 0..1: timetable;

% Solve
solve :: int_search(
    [timetable[c, p] | c in 1..courses, p in 1..periods],
    most_constrained,
    indomain_split,
    complete
) satisfy;

% Constraints
constraint
    % 1. Conflicts: Courses with common students must not be scheduled at the same time
    forall(c1, c2 in 1..courses where c1 < c2) (
        if conflict[c1, c2] = 1 then
            forall(p in 1..periods) (
                timetable[c1, p] + timetable[c2, p] <= 1
            )
        else
            true
        endif
    )
    % 2. Availabilities: Courses can only be scheduled in available periods
    /\
    forall(c in 1..courses, p in 1..periods) (
        if available[c, p] = 0 then
            timetable[c, p] = 0
        else
            true
        endif
    )
    % 3. Rooms: At most `rooms` lectures can be scheduled per period
    /\
    forall(p in 1..periods) (
        sum([timetable[c, p] | c in 1..courses]) <= rooms
    )
    % 4. Number of lectures per course must match the requirement
    /\
    forall(c in 1..courses) (
        sum([timetable[c, p] | p in 1..periods]) = requirement[c]
    );
\end{lstlisting}

\end{tcolorbox}}
    \caption{An example \textsc{MiniZinc} model.}
    \label{fig:minizinc}
\end{figure}

\twocolumn

\begin{figure}[t]
    \centering
    {% Define colors
\definecolor{promptbg}{RGB}{220,240,255}
\definecolor{outputbg}{RGB}{255,250,220}
\definecolor{bordercolor}{RGB}{100,100,100}
\definecolor{headerbg}{RGB}{150,180,220}
\definecolor{headerfg}{RGB}{255,255,255}

\lstset{
    basicstyle=\ttfamily\small,
    columns=fullflexible,
    breaklines=true
}

% Custom Box Style
\tcbset{
    mybox/.style={
        colframe=bordercolor,
        colback=promptbg,
        coltitle=headerfg,
        fonttitle=\bfseries,
        boxrule=0.5mm,
        width=9cm,
        rounded corners,
        enhanced,
        colbacktitle=headerbg,
        title style={left color=headerbg, right color=headerbg, rounded corners},
    },
    outputbox/.style={
        colframe=bordercolor,
        colback=outputbg,
        coltitle=headerfg,
        fonttitle=\bfseries,
        boxrule=0.5mm,
        width=8cm,
        rounded corners,
        enhanced,
        colbacktitle=headerbg,
        title style={left color=headerbg, right color=headerbg, rounded corners},
    }
}

\tcbset{
    mybox2/.style={
        colframe=bordercolor,
        colback=promptbg,
        coltitle=headerfg,
        fonttitle=\bfseries,
        boxrule=0.5mm,
        width=16cm,
        rounded corners,
        enhanced,
        colbacktitle=headerbg,
        title style={left color=headerbg, right color=headerbg, rounded corners},
    },
    outputbox/.style={
        colframe=bordercolor,
        colback=outputbg,
        coltitle=headerfg,
        fonttitle=\bfseries,
        boxrule=0.5mm,
        width=16cm,
        rounded corners,
        enhanced,
        colbacktitle=headerbg,
        title style={left color=headerbg, right color=headerbg, rounded corners},
    }
}

\lstset{
    basicstyle=\ttfamily\small,
    columns=fullflexible,
    breaklines=true,
    tabsize=2, % Sets tab width to 2 spaces
}

\begin{tcolorbox}[mybox, title=data.dzn]

    \begin{lstlisting}
int: courses = 5;
int: periods = 20;
int: rooms = 2;
array[1..courses, 1..periods] of int: available = array2d(1..courses, 1..periods, [
    %  1  2  3  4  5  6  7  8  9  0  1  2  3  4  5  6  7  8  9  0
       0, 0, 1, 1, 1, 1, 1, 1, 1, 1, 1, 1, 0, 1, 1, 0, 1, 1, 1, 1,
       1, 1, 0, 0, 1, 0, 1, 1, 0, 1, 1, 1, 1, 1, 1, 1, 1, 1, 1, 1,
       0, 0, 0, 1, 1, 1, 1, 0, 1, 1, 1, 1, 0, 1, 1, 1, 1, 0, 1, 1,
       1, 1, 1, 0, 0, 0, 1, 1, 1, 1, 1, 1, 1, 1, 1, 1, 1, 0, 1, 1,
       1, 1, 1, 1, 1, 1, 1, 1, 1, 1, 1, 1, 1, 1, 1, 1, 1, 1, 1, 1
]);
array[1..courses, 1..courses] of int: conflict = array2d(1..courses, 1..courses, [
    % Conflict matrix
    0, 1, 0, 0, 1,
    1, 0, 0, 1, 0,
    0, 0, 0, 0, 1,
    0, 1, 0, 0, 1,
    1, 0, 1, 1, 0
]);
array[1..courses] of int: requirement = [6, 10, 14, 6, 4];


\end{lstlisting}

\end{tcolorbox}}
    \caption{An example data instance.}
    \label{fig:data}
    \vspace{-0.4cm}
\end{figure}

% \medskip
\noindent \textbf{Data:} A DZN file representing a concrete problem instance. This instance data is used to validate the code generated by the language model in conjunction with the ground truth output labels. Figure~\ref{fig:data} shows an example data file.

\begin{figure}[t]
    \centering
    % \scalebox{0.93}
    {% Define colors
\definecolor{promptbg}{RGB}{220,240,255}
\definecolor{outputbg}{RGB}{255,250,220}
\definecolor{bordercolor}{RGB}{100,100,100}
\definecolor{headerbg}{RGB}{150,180,220}
\definecolor{headerfg}{RGB}{255,255,255}

\lstset{
    basicstyle=\ttfamily\small,
    columns=fullflexible,
    breaklines=true
}

% Custom Box Style
\tcbset{
    mybox/.style={
        colframe=bordercolor,
        colback=promptbg,
        coltitle=headerfg,
        fonttitle=\bfseries,
        boxrule=0.5mm,
        width=9cm,
        rounded corners,
        enhanced,
        colbacktitle=headerbg,
        title style={left color=headerbg, right color=headerbg, rounded corners},
    },
    outputbox/.style={
        colframe=bordercolor,
        colback=outputbg,
        coltitle=headerfg,
        fonttitle=\bfseries,
        boxrule=0.5mm,
        width=8cm,
        rounded corners,
        enhanced,
        colbacktitle=headerbg,
        title style={left color=headerbg, right color=headerbg, rounded corners},
    }
}

\tcbset{
    mybox2/.style={
        colframe=bordercolor,
        colback=promptbg,
        coltitle=headerfg,
        fonttitle=\bfseries,
        boxrule=0.5mm,
        width=16cm,
        rounded corners,
        enhanced,
        colbacktitle=headerbg,
        title style={left color=headerbg, right color=headerbg, rounded corners},
    },
    outputbox/.style={
        colframe=bordercolor,
        colback=outputbg,
        coltitle=headerfg,
        fonttitle=\bfseries,
        boxrule=0.5mm,
        width=16cm,
        rounded corners,
        enhanced,
        colbacktitle=headerbg,
        title style={left color=headerbg, right color=headerbg, rounded corners},
    }
}

\lstset{
    basicstyle=\ttfamily\small,
    columns=fullflexible,
    breaklines=true,
    tabsize=2, % Sets tab width to 2 spaces
}

\begin{tcolorbox}[mybox, title=output.json]
    \begin{lstlisting}
{
"timetable": [
    [0, 0, 1, 1, 0, 1, 0, 1, 0, 0, 0, 1, 0, 1, 0, 0, 0, 0, 0, 0],
    [1, 1, 0, 0, 1, 0, 0, 0, 0, 0, 0, 0, 1, 0, 1, 1, 1, 1, 1, 1],
    [0, 0, 0, 1, 1, 1, 1, 0, 1, 1, 1, 1, 0, 1, 1, 1, 1, 0, 1, 1],
    [0, 0, 1, 0, 0, 0, 1, 1, 1, 1, 1, 0, 0, 0, 0, 0, 0, 0, 0, 0],
    [1, 1, 0, 0, 0, 0, 0, 0, 0, 0, 0, 0, 1, 0, 0, 0, 0, 1, 0, 0]
  ]
  }

    \end{lstlisting}
\end{tcolorbox}}
    \caption{An example of output of model execution.}
    \label{fig:output}
    \vspace{-0.4cm}
\end{figure}

\medskip
\noindent \textbf{Output:} A JSON file containing the assignments of variables specified in the objective values for a problem. Two essential elements of the output are the final variable configurations (serving as sample solutions for satisfaction problems) and the optimal value (for optimization problems). Figure~\ref{fig:output} shows an example output file.

\subsection{Dataset Statistics} 
\begin{table}[b]
    \centering
    \begin{tabular}{lccc}
        \toprule
        & \# LP & \# MIP & \# CP \\ 
        \midrule
        NLP4LP     &  54 &  13&  0\\
        LPWP       &  1101 &  0&  0\\
        ComplexOR  &  25 &  12&  0\\
        \textsc{Text2Zinc} &  64 & 31 & 15 \\
        \bottomrule
    \end{tabular}
    \caption{Distribution of Problem Types Across Datasets}
    \label{tab:comparison}
\end{table}
\textsc{Text2Zinc} is a cross-domain dataset and the first to encompass optimization and satisfaction problems. 

Table~\ref{tab:comparison} shows the distribution of problem instances modeled with LP, MIP, and CP across datasets.

\subsection{Dataset Verification} Input parameters are validated against DZN files through automated verification. The output files for instances with model files contain the complete \textsc{MiniZinc} model results in JSON format, while instances without models include only the objective value. Model files are provided for all satisfaction problems, and all included models have been verified for successful compilation.

\textsc{Text2Zinc} dataset is actively evolving and expanding with subsequent releases. The initial version includes 45 \textsc{MiniZinc} model files, with plans for additional models in the following updates. All constraint satisfaction problems have been verified to produce valid solutions.

\section{Experimental Methodology}
\label{sec:methodology}

Given \textsc{Text2Zinc}, we present a methodology for using it to test the performance of state-of-the-art LLMs for automated modeling. Our main goal is to establish reasonable baseline performance to serve as a lower bound for emerging approaches.

\smallskip
The task of \textsc{MiniZinc} model generation from natural language can be formally defined with the following Formulation and Evaluation.

\subsection{Problem Formulation}

Let $Input = (description, parameters, output, \\metadata, data)$ represent the input specification where:
\begin{itemize}
    \item $description$ is the natural language description of the problem
    \item $parameters = \{p_1, ..., p_n\}$ is the set of input parameters where each $p_i = (d_i, s_i, \mathbf{h}_i)$ consists of:
        \begin{itemize}
            \item $d_i$: is the natural language definition of the parameter
            \item $s_i$: parameter symbol (e.g., 'num\_warehouses', 'num\_customers', 'P')
            \item $\mathbf{h}_i$: is the shape information of the parameter
        \end{itemize}
    \item $output = \{o_1, ..., o_k\}$ is the set of output variables where each $o_i = (d_i, s_i, \mathbf{h}_i)$ consists of:
        \begin{itemize}
            \item $d_i$: is the natural language definition of the output
            \item $s_i$: output symbol (e.g., 'X', 'Y')
            \item $\mathbf{h}_i$: is the shape information of the output
        \end{itemize}
    \item $metadata$: metadata containing problem properties (e.g., domain, objective type, constraint types)
\end{itemize}

Let $Data$ be a concrete data instance specified in a format like a .dzn file.
The objective is to learn a function $f: (Input, Data) \rightarrow \mathcal{M}$ where $\mathcal{M}$ represents the space of valid \textsc{MiniZinc} models. Given the input $Input$ and data instance $data$, the function should generate a data-compatible model and correctly implement the given specifications.

\subsection{Evaluation}
Given a dataset of $N$ problems, each with a ground truth model $m^*_i$ and generated model $m_i$, we define two key metrics:

\begin{enumerate}
    \item \textbf{Execution Accuracy} ($E_{acc}$): Measures the proportion of generated models that successfully compile and execute:
    \begin{equation}
        E_{acc} = \frac{1}{N}\sum_{i=1}^N \mathbb{I}(execute(m_i) = \text{success})
        \label{eq:execution_accuracy}
    \end{equation}
    where $\mathbb{I}(\cdot)$ is the indicator function.

    \item \textbf{Solution Accuracy} ($S_{acc}$): Measures the proportion of models that achieve the same objective value as the ground truth:
    \begin{equation}
    S_{acc} = \frac{1}{N}\sum_{i=1}^N \mathbb{I}(obj(m_i) = obj(m^*_i))
    \label{eq:solution_accuracy}
    \end{equation}
    where $obj(\cdot)$ returns the objective value and $m^*_i$ is the ground truth model for instance $i$.
\end{enumerate}

\subsection{Solution Approaches}
To explore the effectiveness of different prompting approaches, we conducted experiments on the NLP4LP section of our dataset, comprising 63 optimization problems. Our investigation is organized around three main approaches, each progressively building upon the insights from previous approaches. Based on these approaches, Algorithm~\ref{alg:model-generation} presents a high-level architecture flow of our LLM-assisted model generation pipeline. 

% \begin{enumerate}[label=\textbf{(\alph*)}, leftmargin=*]
\medskip
\noindent \textbf{Vanilla Prompting:} Vanilla prompting involves directly providing the task description and input to the language model without any additional instructions about reasoning steps, problem decomposition, or how to approach the solution.

    \begin{enumerate}[label=\arabic*., leftmargin=*]
     \item \textbf{Basic}: Baseline approach exposing the LLM to:
        \begin{itemize}
            \item Problem description
            \item Expected input parameters
        \end{itemize}
    \item \textbf{Basic + Data Nomenclature \& Examples}: Enhances baseline approach with:
        \begin{itemize}
            \item Parameter definitions
            \item Concrete usage examples sampled from the data instance
        \end{itemize}
    \item \textbf{Basic + Shape Information}: Further augments with:
        \begin{itemize}
            \item Parameter dimensionality information
        \end{itemize}
        
        \item \textbf{Basic + Knowledge Graph}: Further integrates:
        \begin{itemize}
            \item Intermediate knowledge representation
        \end{itemize}
    \end{enumerate}

\medskip
\noindent  \textbf{Chain-of-Thought (CoT) Approaches:} Building upon vanilla prompting, Chain-of-Thought~\cite{chainofthought} prompting adds explicit instructions for the language model to show its reasoning steps, explain its thought process, and documents how it approaches the solution.
    \begin{enumerate}[label=\arabic*., leftmargin=*]
        \item \textbf{CoT}: incorporates reasoning and data nomenclature
        \begin{itemize}
            \item Step-by-step reasoning
            \item Parameter definitions
        \end{itemize}
        \item \textbf{CoT + Examples}: Augments with:
        \begin{itemize}
            \item Concrete usage examples sampled from dzn
        \end{itemize}
        \item \textbf{CoT + Shape Information}: Further augments with:
        \begin{itemize}
            \item Parameter dimensionality information
        \end{itemize}
    \end{enumerate}

 \begin{algorithm}
\caption{LLM-based \textsc{MiniZinc} Model Generation}
\label{alg:model-generation}
\begin{algorithmic}[1]
\Require
    \State $Input$ \Comment{Input specification from input.json}
    \State $Data$ \Comment{Data instance from data.dzn}
    \State $LLM$ \Comment{LLM to prompt}
\State \textbf{Solution Approaches:}
\State vanilla $\leftarrow$ \{
    \State \hspace{1em} Basic,
    \State \hspace{1em}+ Data Nomenclature \& examples,
    \State \hspace{1em}+ Shape information,
    \State \hspace{1em}+ Knowledge Graph information
\State \}

\State cot $\leftarrow$ \{
    \State \hspace{1em}CoT + Data Nomenclature,
    \State \hspace{1em}+ Examples,
    \State \hspace{1em}+ Shape information.
\State \}

\State compositional $\leftarrow$ \{
    \State \hspace{1em}Multi-Call + Composition
\State \}

\State all\_approaches $\leftarrow$ vanilla $\cup$ cot $\cup$ compositional

\For{$approach \in$ all\_approaches}
    \State prompt $\leftarrow$ ConstructPrompt($Input, approach$)
    \State minizinc\_code $\leftarrow$ CallLLM($LLM$, prompt)
    \State SaveToFile(minizinc\_code, $approach$)
\EndFor

\Function{ConstructPrompt}{$Input, approach$}
    \State \Return SampleRelevantInfo($Input, approach$) \Comment{Sample based on strategy}
\EndFunction
\end{algorithmic}
\end{algorithm}

\medskip
\noindent  \textbf{Compositional Approach:} Building upon vanilla and Chain-of-Thought prompting, the compositional approach breaks down the task into smaller, sequential sub-tasks. These sub-tasks generate the parameters and decision variables, constraints, and an objective function. The output for each sub-task is generated independently, and the results are then combined to form the complete solution.
    \begin{enumerate}[label=\arabic*., leftmargin=*]
        \item \textbf{Multi-Call + Composition}: Implements:
        \begin{itemize}
            \item Decomposition into sequential sub-tasks
            \item Followed by combining results together
        \end{itemize}
    \end{enumerate}
% \end{enumerate}

Overall, each solution approach builds upon its predecessors, methodically incorporating additional information and structural elements to improve model generation accuracy. The progression from vanilla approaches to more sophisticated methods allows us to isolate the impact of different prompting elements and identify the most crucial components for successful model generation. We document all our prompts in Appendix~\ref{sec:appendix_prompts} for further visibility.

\subsection{Experimental Setup}
Using language models, our experimental setup systematically evaluates different prompting strategies for model generation. As described in the previous section, we implement three solution approaches: vanilla prompting with progressive enhancements from basic to integration with knowledge graphs), Chain-of-Thought variants with increasing contextual information and a compositional approach using multiple sequential tasks and combining their results. We use the model generation pipeline as shown in Algorithm~\ref{alg:model-generation}.  

We use GPT 4.0 in our experiments on NLP4LP problems from our dataset, which consists of 66 problems. We excluded problems 9, 26, and 27 due to their nature and missing data, resulting in a sample of 63 problems. 

For each input specification $\mathcal{I}$, we generate \textsc{MiniZinc} models using each prompting strategy, where the prompt construction selectively samples relevant information based on the strategy's requirements. The generated models are then evaluated on two metrics: execution accuracy~\ref{eq:execution_accuracy} and solution accuracy~\ref{eq:solution_accuracy}.

\begin{table}[t]
\renewcommand{\arraystretch}{1.3}
\centering
\begin{tabular}{p{4.2cm}cc}
    \toprule
    \textbf{Prompting Method} & \textbf{\makecell[c]{Execution\\Accuracy}} & \textbf{\makecell[c]{Solution\\Accuracy}} \\
    \midrule
    \multicolumn{3}{l}{Vanilla Prompting} \\
    \quad Basic & 0.1904 & 0.0634 \\
    \quad + Data \& examples & 0.3650 & 0.1904 \\
    \quad + Shape & 0.1904 & 0.1269 \\
    \quad + Knowledge Graph  & 0.3492 & 0.1111 \\
    \addlinespace[0.5em]
    \multicolumn{3}{l}{Chain-of-Thought (CoT)} \\
    \quad CoT with data & 0.4285 & 0.1746 \\
    \quad + Examples & \textbf{0.5873} & \textbf{0.2539} \\
    \quad + Shape & 0.5555 & 0.2063 \\
    \addlinespace[0.5em]
    \multicolumn{3}{l}{Compositional} \\
    \quad Multi-Call + Composition & \textbf{0.6031} & 0.2222 \\
    \bottomrule
\end{tabular}
\caption{Experimental results to compare execution and solution accuracy of \textsc{Text2Zinc} baseline approaches.}
\label{tab:experimental-results}
\end{table}
\section{Results and Discussion}
\label{sec:results}

Table~\ref{tab:experimental-results} presents our results which are discussed next. In addition, we release a Hugging Face Leaderboard to benchmark approaches on \textsc{Text2Zinc}\footnote{\url{https://huggingface.co/spaces/skadio/text2zinc-leaderboard}} and welcome submissions on new solution approaches. 
\subsection{Vanilla Approaches}
Our analysis of vanilla prompting approaches reveals several important insights:

% \begin{itemize}
\medskip
    \noindent \textbf{The current limitations of LLMs:} The basic approach achieves only 19.04\% execution accuracy, highlighting that merely exposing the LLM to problem descriptions and parameters is insufficient. The even lower solution accuracy (6.34\%) suggests that successfully executed models often fail to capture the underlying modeling logic correctly.
    
    \medskip
    \noindent \textbf{The value of data context:} Adding parameter definitions and examples significantly improves both metrics (36.50\% execution, 19.04\% solution), nearly doubling performance. This suggests that LLMs benefit from concrete examples to interpret parameters and their data types better.
    
    \medskip
    \noindent \textbf{The necessity of shape information:} Surprisingly, adding shape information degrades performance to baseline levels (19.04\% execution). This counter-intuitive result suggests that additional structural information might overwhelm the model or lead to over-specification.
    
    \medskip
    \noindent \textbf{Knowledge graph integration:} While the KG maintains improved execution accuracy (34.92\%), it leads to poorer solution accuracy (11.11\%). This indicates that structured intermediate knowledge representations alone do not necessarily improve the quality of generated solutions.
% \end{itemize}

\subsection{Chain-of-Thought Approaches}
The Chain-of-Thought approaches demonstrate a clear progression in capability. 

\medskip \noindent \textbf{Baseline CoT:} Even the baseline CoT with data nomenclature outperforms all vanilla approaches (42.85\% execution), suggesting that step-by-step reasoning is fundamental for constraint programming tasks.

\medskip \noindent \textbf{The effect of examples:} The combination of CoT with examples produces the best solution accuracy (25.39\%) and near-best execution accuracy (58.73\%), indicating that step-by-step reasoning also benefits from concrete examples.

\medskip \noindent \textbf{The necessity of shape:} Similar to vanilla approaches, adding shape information to CoT does not improve performance, suggesting an upper limit to beneficial context.

\subsection{Compositional Approach}
Breaking down problems into sequential sub-tasks and stitching their outputs together reveals the value of decomposition:

\medskip \noindent \textbf{Execution vs. Accuracy Trade-off:} Achieving the highest execution accuracy (60.31\%), this approach demonstrates that breaking down model generation into sub-tasks improves the likelihood of producing syntactically valid code. Despite best-in-class execution accuracy, solution accuracy (22.22\%) falls short of the best CoT approach, suggesting that component integration might introduce optimization inefficiencies, either during individual sub-task generation or via stitching.

\medskip

When considering general observations across all approaches, the following patterns emerge: 
    
\smallskip \noindent \textbf{Execution-Solution gap:} Consistently lower solution accuracies (typically 30-50\% of execution accuracies) across different strategies suggest how complex the problem of generating accurate code is. In many cases, we noticed that LLMs occasionally misinterpret constraint optimization problems as satisfiability problems. This fundamental misclassification suggests gaps in LLMs' exposure to fundamental combinatorial concepts.

\smallskip \noindent \textbf{Execution issues:} In our observation, syntax errors are the primary cause of execution failures. This can be attributed to the LLM's limited training on \textsc{MiniZinc}'s specialized syntax and constraints, which differ significantly from widely-used programming languages such as Python. Appendix~\ref{appendix:error_analysis} documents a detailed breakdown of these errors for reference.

\smallskip \noindent \textbf{Information sweet spot:}  Both too little (basic vanilla) and too much (adding shape details) information can be detrimental to performance, suggesting a balanced approach for in-context information. 

\smallskip \noindent \textbf{Reasoning vs. Structure:} The superior performance of CoT and compositional approaches can indicate that how information is processed matters more than the quantity of information provided.

\section{Conclusion}
\label{sec:conclusion}
In this paper, we contribute \textsc{Text2Zinc}, a cross-domain Dataset for modeling optimization and satisfaction problems in \textsc{MiniZinc}. We go beyond the previous attempts by integrating \textit{both} satisfaction and optimization problems within a \textit{unified dataset} using a \textit{solver-agnostic} modeling language.
Our initial findings suggest that LLMs are not yet a plug-and-play solution for combinatorial modeling despite their impressive capabilities in various domains. The consistently low solution accuracy rates across different prompting strategies highlight the inherent challenges in translating natural language specifications into executable \textsc{MiniZinc} models. Nevertheless, the improved performance achieved through Chain-of-Thought reasoning and compositional approaches offers promising directions. 

We emphasize that advancing modeling co-pilots heavily depends on large-scale, high-quality datasets. We are grateful to the efforts of previous work that inspired and help shape \textsc{Text2Zinc}. We encourage the research community to contribute to this and similar efforts. Our experiments with intermediate representations, particularly Knowledge Graphs, reveal that the path from natural language to executable models is not straightforward. While such representations show potential, more research is needed to understand their best utilization. Future work should explore alternative intermediate representations, including named entities, semantic graphs, and agentic frameworks, which might better capture the nuances of formulating combinatorial problems.

\bibliography{main}

\begin{thebibliography}{25}
\providecommand{\natexlab}[1]{#1}

\bibitem[{AhmadiTeshnizi et~al.(2024)AhmadiTeshnizi, Gao, Brunborg, Talaei, and Udell}]{optimus}
AhmadiTeshnizi, A.; Gao, W.; Brunborg, H.; Talaei, S.; and Udell, M. 2024.
\newblock OptiMUS-0.3: Using Large Language Models to Model and Solve Optimization Problems at Scale.
\newblock arXiv:2407.19633.

\bibitem[{Ahrabian et~al.(2023)Ahrabian, Du, Myloth, Ananthan, and Pujara}]{pubgraph}
Ahrabian, K.; Du, X.; Myloth, R.~D.; Ananthan, A. B.~S.; and Pujara, J. 2023.
\newblock PubGraph: A Large-Scale Scientific Knowledge Graph.
\newblock arXiv:2302.02231.

\bibitem[{Bizer et~al.(2009)Bizer, Lehmann, Kobilarov, Auer, Becker, Cyganiak, and Hellmann}]{dbpedia}
Bizer, C.; Lehmann, J.; Kobilarov, G.; Auer, S.; Becker, C.; Cyganiak, R.; and Hellmann, S. 2009.
\newblock DBpedia - A crystallization point for the Web of Data.
\newblock \emph{Journal of Web Semantics}, 7(3): 154--165.
\newblock The Web of Data.

\bibitem[{Bussieck and Meeraus(2004)}]{Bussieck2004}
Bussieck, M.~R.; and Meeraus, A. 2004.
\newblock \emph{General Algebraic Modeling System (GAMS)}, 137--157.
\newblock Boston, MA: Springer US.
\newblock ISBN 978-1-4613-0215-5.

\bibitem[{Chen, Constante-Flores, and Li(2023)}]{optichat}
Chen, H.; Constante-Flores, G.~E.; and Li, C. 2023.
\newblock Diagnosing Infeasible Optimization Problems Using Large Language Models.
\newblock arXiv:2308.12923.

\bibitem[{Dakle et~al.(2023)Dakle, Kad{\i}o{\u{g}}lu, Uppuluri, Politi, Raghavan, Rallabandi, and Srinivasamurthy}]{ner4opt2023}
Dakle, P.~P.; Kad{\i}o{\u{g}}lu, S.; Uppuluri, K.; Politi, R.; Raghavan, P.; Rallabandi, S.; and Srinivasamurthy, R. 2023.
\newblock Ner4opt: Named entity recognition for optimization modelling from natural language.
\newblock In \emph{International Conference on Integration of Constraint Programming, Artificial Intelligence, and Operations Research}, 299--319. Springer.

\bibitem[{Guns(2019)}]{guns2019increasing}
Guns, T. 2019.
\newblock Increasing modeling language convenience with a universal n-dimensional array, CPpy as python-embedded example.
\newblock In \emph{Proceedings of the 18th workshop on Constraint Modelling and Reformulation at CP (Modref 2019)}, volume~19.

\bibitem[{Hooker and van Hoeve(2018)}]{CPandOR}
Hooker, J.; and van Hoeve, W.-J. 2018.
\newblock Constraint programming and operations research.
\newblock \emph{Constraints}, 23.

\bibitem[{Huang et~al.(2024)Huang, Tang, Ge, Hu, Jiang, Wang, Wang, and Zheng}]{orlm}
Huang, C.; Tang, Z.; Ge, D.; Hu, S.; Jiang, R.; Wang, B.; Wang, Z.; and Zheng, X. 2024.
\newblock ORLM: A Customizable Framework in Training Large Models for Automated Optimization Modeling.
\newblock arXiv:2405.17743.

\bibitem[{Jefferson et~al.(1999)Jefferson, Miguel, Hnich, Walsh, and Gent}]{csplib}
Jefferson, C.; Miguel, I.; Hnich, B.; Walsh, T.; and Gent, I.~P. 1999.
\newblock {CSPLib}: A problem library for constraints.

\bibitem[{Kad{\i}o{\u{g}}lu et~al.(2024)Kad{\i}o{\u{g}}lu, Pravin~Dakle, Uppuluri, Politi, Raghavan, Rallabandi, and Srinivasamurthy}]{ner4opt2024}
Kad{\i}o{\u{g}}lu, S.; Pravin~Dakle, P.; Uppuluri, K.; Politi, R.; Raghavan, P.; Rallabandi, S.; and Srinivasamurthy, R. 2024.
\newblock Ner4Opt: named entity recognition for optimization modelling from natural language.
\newblock \emph{Constraints}, 1--39.

\bibitem[{Kjellerstrand(2025)}]{hakank}
Kjellerstrand, H. 2025.
\newblock Hakank's Page.
\newblock \url{https://github.com/hakank/hakank}.
\newblock GitHub repository, Retrieved February 13th, 2025.

\bibitem[{Li et~al.(2023)Li, Mellou, Zhang, Pathuri, and Menache}]{optiguide}
Li, B.; Mellou, K.; Zhang, B.; Pathuri, J.; and Menache, I. 2023.
\newblock Large Language Models for Supply Chain Optimization.
\newblock arXiv:2307.03875.

\bibitem[{Michailidis, Tsouros, and Guns(2024)}]{michailidis_et_al:LIPIcs.CP.2024.20}
Michailidis, K.; Tsouros, D.; and Guns, T. 2024.
\newblock {Constraint Modelling with LLMs Using In-Context Learning}.
\newblock In Shaw, P., ed., \emph{30th International Conference on Principles and Practice of Constraint Programming (CP 2024)}, volume 307 of \emph{Leibniz International Proceedings in Informatics (LIPIcs)}, 20:1--20:27. Dagstuhl, Germany: Schloss Dagstuhl -- Leibniz-Zentrum f{\"u}r Informatik.
\newblock ISBN 978-3-95977-336-2.

\bibitem[{Nethercote et~al.(2007)Nethercote, Stuckey, Becket, Brand, Duck, and Tack}]{minizinc}
Nethercote, N.; Stuckey, P.~J.; Becket, R.; Brand, S.; Duck, G.~J.; and Tack, G. 2007.
\newblock MiniZinc: Towards a Standard CP Modelling Language.
\newblock In Bessi{\`e}re, C., ed., \emph{Principles and Practice of Constraint Programming -- CP 2007}, 529--543. Berlin, Heidelberg: Springer Berlin Heidelberg.
\newblock ISBN 978-3-540-74970-7.

\bibitem[{Prasath and Karande(2023)}]{synthesis}
Prasath, G.; and Karande, S. 2023.
\newblock Synthesis of Mathematical programs from Natural Language Specifications.
\newblock arXiv:2304.03287.

\bibitem[{Ramamonjison et~al.(2022{\natexlab{a}})Ramamonjison, Li, Yu, He, Rengan, Banitalebi-Dehkordi, Zhou, and Zhang}]{nl4opt}
Ramamonjison, R.; Li, H.; Yu, T.~T.; He, S.; Rengan, V.; Banitalebi-Dehkordi, A.; Zhou, Z.; and Zhang, Y. 2022{\natexlab{a}}.
\newblock Augmenting Operations Research with Auto-Formulation of Optimization Models from Problem Descriptions.

\bibitem[{Ramamonjison et~al.(2022{\natexlab{b}})Ramamonjison, Yu, Li, Li, Carenini, Ghaddar, He, Mostajabdaveh, Banitalebi-Dehkordi, Zhou, and Zhang}]{pmlr-v220-ramamonjison23a}
Ramamonjison, R.; Yu, T.; Li, R.; Li, H.; Carenini, G.; Ghaddar, B.; He, S.; Mostajabdaveh, M.; Banitalebi-Dehkordi, A.; Zhou, Z.; and Zhang, Y. 2022{\natexlab{b}}.
\newblock NL4Opt Competition: Formulating Optimization Problems Based on Their Natural Language Descriptions.
\newblock In Ciccone, M.; Stolovitzky, G.; and Albrecht, J., eds., \emph{Proceedings of the NeurIPS 2022 Competitions Track}, volume 220 of \emph{Proceedings of Machine Learning Research}, 189--203. PMLR.

\bibitem[{Singhal(2012)}]{knowledgegraph}
Singhal, A. 2012.
\newblock Introducing the Knowledge Graph: Things, not strings.
\newblock \url{https://www.blog.google/products/search/introducing-knowledge-graph-things-not/}.
\newblock Google Blog.

\bibitem[{Suchanek, Kasneci, and Weikum(2008)}]{yago}
Suchanek, F.~M.; Kasneci, G.; and Weikum, G. 2008.
\newblock YAGO: A Large Ontology from Wikipedia and WordNet.
\newblock \emph{Journal of Web Semantics}, 6(3): 203--217.
\newblock World Wide Web Conference 2007Semantic Web Track.

\bibitem[{Tsouros et~al.(2023)Tsouros, Verhaeghe, Kadioglu, and Guns}]{tsouros2023holy}
Tsouros, D.; Verhaeghe, H.; Kadioglu, S.; and Guns, T. 2023.
\newblock Holy Grail 2.0: From Natural Language to Constraint Models.
\newblock \emph{arXiv preprint arXiv:2308.01589}.

\bibitem[{van Hoeve and Katriel(2006)}]{globalconstraints}
van Hoeve, W.-J.; and Katriel, I. 2006.
\newblock \emph{Global constraints}.
\newblock "Elsevier".

\bibitem[{Voboril, Ramaswamy, and Szeider(2024)}]{streamllm}
Voboril, F.; Ramaswamy, V.~P.; and Szeider, S. 2024.
\newblock Realtime Generation of Streamliners with Large Language Models.
\newblock arXiv:2408.10268.

\bibitem[{Wei et~al.(2023)Wei, Wang, Schuurmans, Bosma, Ichter, Xia, Chi, Le, and Zhou}]{chainofthought}
Wei, J.; Wang, X.; Schuurmans, D.; Bosma, M.; Ichter, B.; Xia, F.; Chi, E.; Le, Q.; and Zhou, D. 2023.
\newblock Chain-of-Thought Prompting Elicits Reasoning in Large Language Models.
\newblock arXiv:2201.11903.

\bibitem[{Xiao et~al.(2023)Xiao, Zhang, Wu, Xu, Wang, Han, Fu, Zhong, Zeng, Song et~al.}]{complexor}
Xiao, Z.; Zhang, D.; Wu, Y.; Xu, L.; Wang, Y.~J.; Han, X.; Fu, X.; Zhong, T.; Zeng, J.; Song, M.; et~al. 2023.
\newblock Chain-of-Experts: When LLMs Meet Complex Operations Research Problems.
\newblock In \emph{The Twelfth International Conference on Learning Representations}.

\end{thebibliography}
\newpage
\begin{appendixformat}

\section{Appendix - GPT-4 MiniZinc Code Generation Prompts}
\label{sec:appendix_prompts}

We present the details of the prompts used in our experiments below.

\subsection{Vanilla Prompting Approach}

\subsubsection{\textsc{Pure Vanilla}}
You are an expert MiniZinc developer.\newline\newline
Generate Minizinc code from a given problem description with additional information about the parameters provided.\newline\newline
The MiniZinc code should assume that the data needed, will be provided in a specific format through a .dzn file, so the generated code should assume the same names defined in the \textbf{input data nomenclature}.\newline\newline
Please do not generate any other token, except the MiniZinc code.\newline\newline
\textbf{Problem Description}:\newline
\{problem\_description\}\newline\newline
\textbf{Input Data Nomenclature}:\newline
\{data\_nomenclature\}

\subsubsection{\textsc{Vanilla \texttt{+} data nomenclature \texttt{+} examples}}
You are an expert MiniZinc developer.\newline\newline
Generate Minizinc code from a given problem description with additional information about the parameters provided.\newline\newline
The MiniZinc code should assume that the data needed, will be provided in a specific format through a .dzn file, so the generated code should assume the same names/data-types defined in the \textbf{input data nomenclature and examples}.\newline\newline
Please do not generate any other token, except the MiniZinc code.\newline\newline
\textbf{Problem Description}:\newline
\{problem\_description\}\newline\newline
\textbf{Input Data Nomenclature and Examples}:\newline
\{data\_nomenclature\}
\subsubsection{\textsc{Vanilla \texttt{+} data nomenclature \texttt{+} examples \texttt{+} kg graph}}
You are an expert MiniZinc developer.\newline\newline
Generate Minizinc code from using the following information:\newline\newline
1. Problem Description: A formal description describing the optimization problem.\newline
2. Knowledge Graph: Detailing Parameters, Variables, Constraints and Objective.\newline
3. Input Data Nomenclature: The MiniZinc code should assume that the data needed, will be provided in a specific format through a .dzn file, so the generated code should assume the same names defined in the \textbf{input data nomenclature}.\newline\newline
Please do not generate any other token, except the MiniZinc code.\newline\newline
\textbf{Problem Description}:\newline
\{problem\_description\}\newline\newline
\textbf{Knowledge Graph}:\newline
```\newline
\{knowledge\_graph\}\newline
```\newline\newline
\textbf{Input Data Nomenclature}:\newline
\{data\_nomenclature\}
\subsection{Chain-of-Thought (CoT) Approach}
\subsubsection{\textsc{chain-of-thought (cot) \texttt{+} data nomenclature \texttt{+} examples}}
You are an expert MiniZinc developer.\newline\newline
Generate Minizinc code from a given problem description with additional information about the parameters provided.\newline\newline
The MiniZinc code should assume that the data needed, will be provided in a specific format through a .dzn file, so the generated code should assume the same names, shapes and data-types defined in the \textbf{input data nomenclature and examples}.\newline\newline
When generating the code, follow this format:\newline\newline
```\newline
\% Parameters\newline\newline
\% Variables\newline\newline
\% Constraints\newline\newline
\% Objective\newline\newline
```\newline\newline
Also, make sure to follow the following principles when generating the code:\newline\newline
\textbf{General Principles}:\newline
1. The generated code should assume that data will be provided via a ".dzn" file. Do not declare values directly from the input data nomenclature and examples within the MiniZinc model.\newline
2. Adhere to the input data nomenclature and examples precisely when declaring input parameter names and their data types.\newline
3. Use bounded variables whenever possible. If bounds are explicit (e.g., non-negative), include them as constraints.\newline
4. When defining arrays of variables, ensure bounds are integers. Apply element-wise constraints in a separate constraint block if bounds depend on array elements.\newline
5. When defining arrays of variables, ensure bounding constraints are applied separately rather than during initialization to avoid type mismatches.\newline
6. Define explicit bounds for all variables used in linear expressions, either in their declaration or through additional constraints.\newline
7. Separate constraints into distinct constraint blocks whenever possible.\newline
8. Use direct and succinct definitions for constraints in the model.\newline
9. Utilize global constraints as much as possible.\newline
10. Declare all parameters and sets before using them in other declarations to avoid circular dependencies and ordering issues.\newline
11. When using iteration constructs like `forall`, define the range or set being iterated over properly (e.g., use `1..M` instead of `M` for iteration).\newline
12. Ensure operands in operations are of compatible types to prevent coercion errors.\newline
13. Declare all identifiers (such as indices or ranges like `n`) before using them in any array or parameter declarations.\newline
14. Ensure type consistency in expressions to avoid coercion errors. Explicitly cast types if necessary.\newline
15. Ensure there is only one objective, which will be a maximization, minimization, or a satisfy problem. Do not forget the `solve` keyword.\newline
Please do not generate any other token, except the MiniZinc code.\newline\newline
\textbf{Problem Description}:\newline
\{problem\_description\}\newline\newline
\textbf{Input Data Nomenclature and Examples}:\newline
\{data\_nomenclature\}

\subsection{Compositional Approach}
\subsubsection{\textsc{multi call prompt \texttt{+} Composition}}
\subsubsection{Parameters and Variables}
You are an expert MiniZinc developer.\newline\newline
Generate MiniZinc code for the Parameters and Variables from a given problem description with additional information about input data provided.\newline\newline
The MiniZinc code should assume that the data needed will be provided in a specific format through a .dzn file, so the generated code should assume the same names/data-types defined in the \textbf{input data nomenclature and examples}.\newline\newline
When generating the code, follow this format:\newline\newline
```minizinc\newline
\% Parameters\newline\newline
\% Variables\newline
```\newline
Also, make sure to follow the following principles when generating the code:\newline\newline
\textbf{General Principles}:\newline
1. The generated code should assume that data will be provided via a ".dzn" file. Do not declare values directly from the input data nomenclature and examples within the MiniZinc model.\newline
2. Adhere to the input data nomenclature and examples precisely when declaring input parameter names and their data types.\newline
3. Use bounded variables whenever possible. If bounds are explicit (e.g., non-negative), include them as constraints.\newline
4. When defining arrays of variables, ensure bounds are integers. Apply element-wise constraints in a separate constraint block if bounds depend on array elements.\newline
5. When defining arrays of variables, ensure bounding constraints are applied separately rather than during initialization to avoid type mismatches.\newline
6. Define explicit bounds for all variables used in linear expressions, either in their declaration or through additional constraints.\newline
7. Declare all parameters and sets before using them in other declarations to avoid circular dependencies and ordering issues.\newline
8. Declare all identifiers (such as indices or ranges like `n`) before using them in any array or parameter declarations.\newline
9. Ensure that all indices and sets used in parameter and variable declarations are declared beforehand.\newline
10. When declaring variables, ensure they are appropriately typed (e.g., `int`, `float`).\newline
11. Variables should have meaningful names related to the problem description.\newline
12. Include comments to briefly describe each parameter and variable for clarity.\newline\newline
\textbf{Problem Description}:\newline
\{problem\_description\}\newline\newline
\textbf{Input Data Nomenclature and Examples}:\newline
\{data\_nomenclature\}
\subsubsection{Constraints}
You are an expert MiniZinc developer.\newline\newline
Generate MiniZinc code for the Constraints from a given problem description with additional information about the parameters provided.\newline\newline
Given the Parameters and Variables part of the code, generate only the constraints.\newline\newline
When generating the code, follow this format:\newline\newline
```minizinc\newline
\% Constraints\newline\newline
```\newline
Also, make sure to follow the following principles when generating the code:\newline\newline
\textbf{General Principles}:\newline
1. Separate constraints into distinct constraint blocks whenever possible.\newline
2. Utilize global constraints as much as possible.\newline
3. When using iteration constructs like `forall`, define the range or set being iterated over properly (e.g., use `1..M` instead of `M` for iteration).\newline
4. Ensure operands in operations are of compatible types to prevent coercion errors.\newline
5. Ensure type consistency in expressions to avoid coercion errors.\newline
6. Clearly comment on the purpose of each constraint for clarity and maintenance.\newline
7. Avoid hardcoding values; use parameters and variables instead.\newline
8. Use meaningful names for all constraint blocks.\newline
9. Only generate constraints and do not generate objective.\newline\newline
\textbf{Problem Description}:\newline
\{problem\_description\}\newline\newline
\textbf{Input Data Nomenclature and Examples}:\newline
\{data\_nomenclature\}\newline\newline
\textbf{Parameters and Variables}:\newline
```minizinc\newline
\{parameters\_and\_variables\}\newline
```
\subsubsection{Objective}
You are an expert MiniZinc developer.\newline\newline
Generate MiniZinc code for the Objective from a given problem description with additional information about the parameters, variables and constraints provided.\newline\newline
Given the Parameters, Variables, and Constraints sections of the code, generate only the objective.\newline\newline
When generating the code, follow this format:\newline\newline
```minizinc\newline
\% Objective\newline\newline
```\newline\newline
Also, make sure to follow the following principles when generating the code:\newline\newline
\textbf{General Principles}:\newline\newline
1. Ensure there is only one objective, which will be a maximization, minimization, or a satisfy problem. Do not forget the `solve` keyword.\newline
2. Ensure the objective function aligns with the problem description.\newline
3. Verify the correct usage of variables, parameters and constraints in the objective function.\newline\newline
\textbf{Problem Description}:\newline
\{problem\_description\}\newline\newline
\textbf{Input Data Nomenclature and Examples}:\newline
\{data\_nomenclature\}\newline\newline
\textbf{Parameters and Variables}:\newline
```minizinc\newline
\{parameters\_and\_variables\}\newline
```\newline\newline
\textbf{Constraints}:\newline
```minizinc\newline
\{constraints\}\newline
```
\subsubsection{Stitch}
You are an expert MiniZinc developer.\newline\newline
Given the Parameters, Variables, Constraints, and Objective sections of the code, stitch them into a complete solution for the optimization problem.\newline\newline
When stitching the code, follow this format:\newline\newline
```minizinc\newline
\% Parameters\newline\newline
\% Variables\newline\newline
\% Constraints\newline\newline
\% Objective\newline\newline
```\newline\newline
Ensure the following principles for syntactic accuracy and logical consistency:\newline\newline
\textbf{General Principles}:\newline
1. Verify that all intermediate sections (parameters, variables, constraints, objective) are consistent and correctly referenced.\newline
2. Confirm that the final MiniZinc code is syntactically accurate and logically coherent.\newline
3. Ensure that the code sections are properly integrated, maintaining the prescribed format.\newline
4. Check for and resolve any circular dependencies or ordering issues in declarations.\newline
5. Check for and resolve any coercion issues.\newline
5. Validate type consistency across all expressions and declarations.\newline
6. Utilize clear and concise comments to describe each section and its components.\newline
7. Make sure global constraints are utilized where applicable to enhance model efficiency.\newline
8. Ensure only one objective is defined, using the `solve` keyword appropriately.\newline\newline
\textbf{Problem Description}:\newline
\{problem\_description\}\newline\newline
\textbf{Input Data Nomenclature and Examples}:\newline
\{data\_nomenclature\}\newline\newline
\textbf{Parameters and Variables}:\newline
```minizinc\newline
\{parameters\_and\_variables\}\newline
```\newline\newline
\textbf{Constraints}:\newline
```minizinc\newline
\{constraints\}\newline
```\newline\newline
\textbf{Objective}:\newline
```minizinc\newline
\{objective\}\newline
```
\newpage
\section{\textsc{knowledge graph generation prompt}}
\label{appendix:knowledge_graph_generation}
Given an optimization problem description and the nomenclature to be used for parameters.\newline\newline
Please generate a knowledge-graph.\newline\newline
\textbf{Problem Description}:\newline
\{problem\_description\}\newline\newline
\textbf{Input Data Nomenclature}:\newline
\{data\_nomenclature\}\newline\newline
\textbf{Steps to Generate a Knowledge Graph (KG)}:\newline\newline
1. \textbf{Identify Parameters}:\newline
   - Extract each parameter individually, specifying its type and name.\newline
   - Ensure that parameter names are aligned with the predefined nomenclature.\newline
   - Record any explicit bounds for parameters, if specified.\newline
   - Validate against the problem description to ensure that all relevant parameters are included, checking for any implicitly stated ones.\newline
2. \textbf{Identify Variables}:\newline
   - Identify each variable, noting their types and names.\newline
   - Align variable names with the nomenclature provided, ensuring consistency.\newline
   - Determine and document any bounds (explicit or inferred) for the variables.\newline
   - Cross-verify with the problem description to confirm all necessary variables are accounted for, including those not explicitly mentioned in the nomenclature.\newline
3. \textbf{Identify Constraints}:\newline
   - Detail each constraint separately, recording its description, formula, and associated variables.\newline
   - Classify constraints based on their nature (e.g., linear, nonlinear, global etc..).\newline
   - Highlight if any constraints are global, impacting multiple variables or conditions across the model.\newline
4. \textbf{Identify Objective}:\newline
   - Clearly define the objective function, including its description, mathematical representation, and the variables it affects.\newline
   - Note the type of optimization (minimization, maximization, satisfy) and any relevant constraints tied to the objective.\newline
   - Ensure the objective aligns with the overall goal of the optimization problem as described.\newline
5. \textbf{Generate Turtle (TTL) Representation}:\newline
   - Construct the Turtle format representation using the gathered information on parameters, variables, constraints, and the objective.\newline
   - Sequentially organize the definitions of parameters, variables, constraints, and the objective within the Turtle file.\newline
   - Avoid generalizing the problem statement; focus on defining each component distinctly as outlined in the steps.\newline\newline
Please do not generate any other token other than the knowledge graph itself. No titles, description or markup necessary.
\newpage
\subsection{Example Knowledge Graph}
\label{appendix:knowledge_graph_example}
\textbf{Unstructured Description}\newline
A firm produces different goods using different raw materials. The firm has a limited amount of each raw material available. Each good requires a specific amount of raw materials to be produced. Each good results in an amount of revenue produced per unit. How much of each good should the firm produce to maximize its total revenue?\newline\newline
\textbf{Knowledge Graph}
\begin{lstlisting}[language=SPARQL, caption=Knowledge Graph in TTL format]
@prefix : <http://example.org/firm_optimization#> .

# Parameters

:M a :Parameter ;
   :name "Number of different goods" ;
   :type "Integer" ;
   :description "Total number of different goods produced by the firm" .

:N a :Parameter ;
   :name "Number of different raw materials" ;
   :type "Integer" ;
   :description "Total number of different raw materials used by the firm" .

:Available a :Parameter ;
   :name "Available amount of raw material i" ;
   :type "1-D Array" ;
   :description "Amount of each raw material available for production" .

:Requirements a :Parameter ;
   :name "Requirement of raw material i to produce one unit of good j" ;
   :type "2-D Array" ;
   :description "Amount of each raw material required to produce one unit of each good" .

:Prices a :Parameter ;
   :name "Revenue from selling one unit of good j" ;
   :type "1-D Array" ;
   :description "Revenue earned by selling one unit of each good" .

# Variables

:UnitsProduced a :Variable ;
   :name "Units produced of good j" ;
   :type "1-D Array" ;
   :bounds "0 to infinity" ;
   :description "Number of units of each good produced" .

# Constraints

:MaterialAvailability a :Constraint ;
   :description "Each good's production must not exceed available raw materials" ;
   :formula "Sum(Requirements[i][j] * UnitsProduced[j] for j in 1..M) <= Available[i] for i in 1..N" .

# Objective

:MaximizeRevenue a :Objective ;
   :description "Maximize the total revenue from selling the goods" ;
   :formula "Sum(Prices[j] * UnitsProduced[j] for j in 1..M)" ;
   :type "Maximization" .
\end{lstlisting}
\newpage
\section{Error Analysis}
\label{appendix:error_analysis}
\begin{table}[h]
\centering
\renewcommand{\arraystretch}{1.0}
\begin{tabular}{l p{10cm}}
\hline
\textbf{Error Type} & \textbf{Commentary} \\
\hline
Syntax Errors & These errors indicate issues with the MiniZinc code syntax. They may include unexpected tokens, missing semicolons, or incorrect use of language constructs. For example: \texttt{`Error: syntax error, unexpected '|''}. \\
\hline
Type Coercion Errors & These errors happen when there is a mismatch between expected and provided data types, such as expecting an integer but receiving a float. Example: \texttt{`Error: type error: cannot determine coercion from type var float to type var int'}. \\
\hline
Undefined Identifiers & These errors are due to the use of variables or identifiers that have not been declared or are out of scope. Example: \texttt{`Error: type error: undefined identifier `i', did you mean `A'?'}. \\
\hline
Array and Indexing Issues & Errors related to improper use of arrays or indexing problems. This can include out-of-bounds errors or incorrect array dimensions. Example: \texttt{`Error: evaluation error: Index set mismatch. Declared index sets of `Benefit' are [1..5,1..3], but is assigned to array with index sets [1..5, 1..2]'}. \\
\hline
Constraint Definition Errors & These errors occur when there is an issue with the constraints defined in the model. This can include logical inconsistencies or incorrect use of constraint functions. Example: \texttt{`Error: solver backend cannot handle constraint: float\_lin\_ne'}. \\
\hline
Model Instantiation Errors & Errors related to the instantiation of the model, such as problems with set definitions or parameter initialization. Example: \texttt{`Error: type error: generator expression must be (par or var) set of int or array, but is `int'}. \\
\hline
Missing Data in DZN Files & These errors occur when the necessary data is not provided in the DZN files. This may lead to missing parameters or sets that are required by the model. Example: \texttt{`Error: type error: variable `K' must be defined (did you forget to specify a data file?)'}. \\
\hline
\end{tabular}
\caption{Categorization of MiniZinc Errors with Commentary.}
\end{table}
\end{appendixformat}
\end{document}